# Biologically inspired design framework for Robot in Dynamic Environments using Framsticks

Raja Mohamed S[1], Dr Raviraj P[2]

[1] Associate Professor / Dept of CSE
Sri Vishnu Engg College for Women, WG District, AP, India
rajamohameds@svecw.edu.in

[2] Professor / Dept of CSE
SKP Engg College, Tiruvannamalai, TN, India
raviraj_it@yahoo.co.in

**Abstract.** Robot design complexity is increasing day by day especially in automated industries. In this paper we propose biologically inspired design framework for robots in dynamic world on the basis of Co-Evolution, Virtual Ecology, Life time learning which are derived from biological creatures. We have created a virtual khepera robot in Framsticks and tested its operational credibility in terms hardware and software components by applying the above suggested techniques. Monitoring complex and non complex behaviors in different environments and obtaining the parameters that influence software and hardware design of the robot that influence anticipated and unanticipated failures, control programs of robot generation are the major concerns of our techniques.

**Keywords:** Biology, Khepera, Framsticks, Framework, Simulation.

## 1  Introduction to Virtual World

The realistic physical modeling of characters in games and virtual worlds is becoming a viable alternative to more traditional animation techniques. Physical modeling of characters and environments can enhance realism and allow users to interact with the virtual world much more freely. By modeling the forces and torques acting on bodies within a virtual environment, detecting and responding to collisions between bodies, respecting constraints arising from, for example, joints in articulated multi-body objects, etc., the system behaves in a believable manner in all situations, and the user can therefore be allowed to interact with it much more freely. The growing popularity of this approach is demonstrated by the appearance over the last couple of years of a number of off-the-shelf physics engines aimed at games programmers [1]. While these engines handle the modeling of inanimate bodies, programmers are still left with the task of writing controllers for motile objects within the environment (e.g. cars, human characters, monsters, etc.). Writing a controller for a physically modeled character is a question of calculating the appropriate forces to apply to each body part at each point of time to achieve the desired movement (e.g. a realistic walking gait for a human character). Artificial life techniques can be useful in automating this task [2] [3]. For example, artificial evolution can generate suitable controllers for simple behaviors, given only a high level description of that behavior in terms of a fitness function. In this paper, the state of the art in evolving controllers, and also in evolving the characters' body shapes, is described. It is then suggested that current approaches will not be able to scale up to more complicated behaviors.

## 2   Background

The design of robotic systems is particularly challenging due to the breadth of engineering expertise required. In general, a robot may be represented schematically as shown in Figure 1a. In brief, the Control box refers to any off-board tele-operator, whether it is a computer or a human. Robots can be created without this component or the following Communication component, which represents whatever means the robot, has to pass and receive information. Presiding over the control of the robot is High-End which takes in objectives from the outside as well as internal objectives, compares those to the information provided by the sensors, and then issues objectives to "Low-End. Examples of high-level objectives are navigational imperatives, or data collection routines, or manipulation directives. The needs of these objectives are translated by high-level control into specific needs for individual actuators, primarily related to motion (position, speed, etc…). Low-level control represents the hardware and software directly responsible for producing the excitation signal to the system actuators. Again, information from the sensors is compared to the desired goal, and the appropriate stimulus is fed to the Control Signal. The actuators, in turn, act on the Mechanics, which is an enclosure name for the system, whose dynamics and statics the controllers seek to modify. In general, the Mechanics refers to the mechanical components of a robot. However, it could just as easily refer to a chemical solution, a magnetic field, or any other of a myriad of other physical systems.

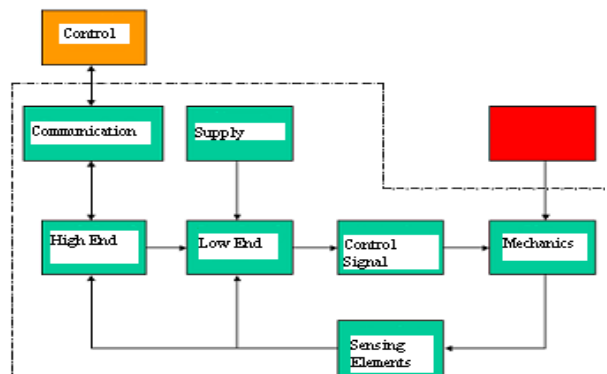

**Figure 1.**  (a) Schematics of General Robot design

Simulation is a technique for exploring interesting regions of this immense landscape of robot design. It is a platform for generating suitable and interesting forms and behaviors, not limited by the preconceptions of a human designer's imagination. It can only get off the ground, if the initial randomly generated robots with a non-zero score on the fitness function. For example, consider an attempt to evolve a behavior whereby a robot needed to process complex visual data about the movement of another robot standing in front of it, and use this to decide whether that object is a friend or foe. This enterprise would clearly have little chance of success if the evolutionary process was starting from scratch.

## 3   Biologically Inspired Techniques

Researchers in the field of evolutionary robotics have considered various methods to overcome the difficulties like selecting a non-zero fitness function, mutation etc. These generally involve either the incremental acquisition of increasingly complicated tasks [3], or the decomposition of the task into a sequence of easier tasks together with some way of combining them. A problem with many of these approaches is that the decomposition of the task into easier and/or incremental steps is something of an art in itself and there are no general guidelines to suggest the most appropriate way to do this sensible task. Decomposition from the designer's point of view may not be the best route by which one can evolve a complex behavior. Despite the problems described above with evolving single robot to perform complex tasks, there are a number of alternative approaches that have shown some signs of success. Five such popular methods inspired from biological creatures are given below:

### 3.1   Co-Evolution

Co-evolution is a very promising technique for developing complex behaviors, especially when there is a competition between two or more robots. The idea is that rather than evolving a robot against a fixed fitness function, two robots are used instead, with one evolving against the other. For example, Hillis evolved efficient number sorting algorithms by co-evolving a population of candidate algorithms against a population of numbers to be sorted [5]. As the sorting algorithms got improved, so are the population of numbers to be sorted, evolved to present tougher challenges to the algorithms.

### 3.2   Virtual Ecologies

Here we can concurrently simulate no of robots of same type having same fitness function against virtual world parameters like survival time, ability to complete the task and so on based on natural selection. In order to identify complex and non complex behaviors some steps have been proposed [7]. Now single physical simulation performed concurrently in parallel mode to test against other robots.

### 3.3 Lifetime Learning

Virtual robot controller gets improved by evolution, but they do not actually adapt whilst an individual robot is being simulated. Some recent results from evolutionary robotics suggest that combining an evolutionary algorithm with the ability of the controllers' to adapt or learn over an individual robot's lifetime can lead to improved robustness and complexity of behaviors compared to evolution by itself [4]. Giving an individual robot the ability to adapt and learn during its lifetime effectively smoothes the search space over which evolution is happening, thereby helping the process to progress. It is reasonable to assume that adding these sorts of abilities to our artificial robot will improve its ability to evolve complex behaviors just as it has done in evolutionary robotics.

### 3.4 Behavioral Primitives

The evolution of complex behaviors can in general be evolved by task decomposition. Rather than trying to evolve complex behaviors, another approach is to evolve a collection of primitive behaviors, and then use other, non-evolutionary techniques for combining these primitive into more complicated sequences. The task of programming a robot using this approach is like commanding it perform an action in the virtual world.

### 3.5 User Guided Evolution

Another alternative to supplying a fixed fitness function to the genetic algorithm is to present the user with a variety of robots from the evolving population at various intervals, and allow them to select their favorite prototypes to be used as the foundation of the next generation. The user may select it under any criteria, and can therefore guide the path of evolution according to their own preferences without having to formally instruct the individual robots.

## 4 Moving Robot Design

Simulation and optimization of digital robot were attempted by several researchers. In this case we used Framsticks software to simulate and evolve digital robots.

### 4.1 Framsticks

Framsticks is 3D simulation software for agents and controllers. It allows using user-defined experiment setups, fitness functions, and neurons (sensor network) and is suitable for testing various research hypotheses with fast 3D simulation and evolutionary optimization. The physical structure of Framsticks agents is made of parts (material points) and joints which have touch, move, and rotation sensors along with the end points of its structure. The control system is made of neurons (including sensors and actuators) and their connections. Framsticks supports multiple genetic representations and operators and it ranges from simple and direct descriptions of agents to the ones encoding developmental process in genotypes. Further possibilities are like performing simulation of several real life creatures and storing in a library to use later.

**4.2   Autonomous Moving Robot**

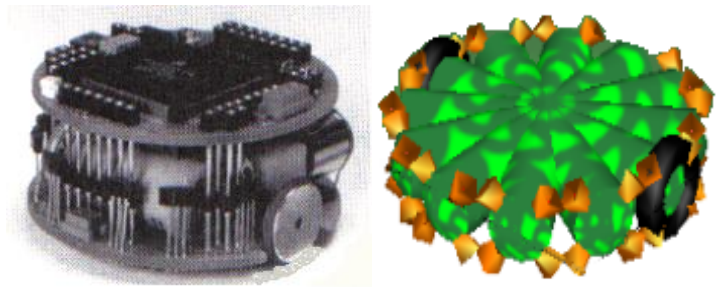

**Figure 1.**  (b) K-Team's Khepera Robot (c) Framsticks Khepera Robot

Khepera robot having 2 wheels (left, right) and 10 sensors in all directions to pick the data from the nearby places and to decide the direction of navigation with out hitting obstacles or reaching a target in the virtual world. Sensor data will be passed to the brain or NN where decision will be made to select the direction based on threshold values of each neuron. Khepera robot is having a neural network associated with it to make decisions based upon real time parameters that are picked up from the environment. Here each node in represents a neuron (i.e.) two neurons for each wheel and a neuron for each sensor as well. Other parameters like ambience, surface type, obstacle sensors all to have its effect over the decision which would be taken by the robot before making a movement as shown in Figure 2 and Figure 3.

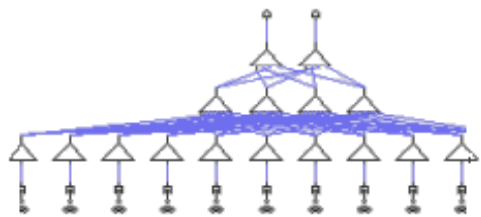

**Figure 2.**  Neural network of Khepera Robot

**4.3   Dynamic Simulation**

The Robot is given by a set of rigid bodies called links and their kinematic structure. The links are described by their physical attributes like inertia properties, mass, center of mass and their geometry in a polygonal representation. The robot's kinematic structure is defined by connecting links with joints. Additional parameters are global gravity and maximum and minimum joint forces and angles. Further forces result from the collision of the robot links with the floor. Collision between robot links is not taken into account in the actual implementation.

Now we are going to design an autonomous moving robot by first creating the same in framsticks' virtual world. After that we will apply several experimental parameters to test the stability of robot in different environments available in framsticks (e.g. flat land, bumpy land) as per the techniques that are proposed earlier and both artificial and real robots are shown in Figure 1b and 1c. Khepera robot's kinematics structure is defined using genome editor exclusively available for framsticks and here we use f1 format with the following genomic sequence as it describes the robot in terms of small segments in framsticks.

**(rrX(lX(llSSSEEX[T:1],),lmXMMMMEEX[|1:2,-1:-3]rrSEEX[T:-0.407](SSIISSLlEEX,,SSIISSLlEEX),))**

After creating this structure we have simulated the above the digital robot in the framsticks virtual world with several environmental conditions with different set of parametric values. Simulation was conducted in a system with configuration of Intel Pentium Dual Core E2140 @ 2.8 GHz and 2 GB RAM in framsticks 3.0.

### 4.3.1 Fitness Function

In order to analyze the behavioral parameters of Khepera robot we need to define a fitness function that can monitor the properties like time taken to perform an action, no of rotations performed by each wheels. Even parallelized function could also be used to save computation time.

$$F(t) = \{\Sigma\ S_i\text{-}T_h \Rightarrow (L_w, R_w) \Rightarrow r\} \qquad (1)$$

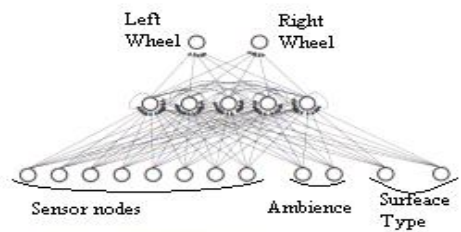

**Figure 3.** Khepera's neural net connection pattern in Framsticks

In the above fitness function **Si** represents signal from all sensor to be compared against **Th** (threshold value) and movement made by left ($L_w$) and right ($R_w$) sided wheels in one unit time or to reach a target from the current location where as **r** denotes no of rotations performed to reach target. This was repeatedly applied for

## 5 Experiments and Results

Khepera robot's fitness value for all types of actions will be calculated from the distance between the robot wheels and the surface by setting default unit value 1. If this condition is not satisfied or violated then a penalty will be applied to over all fitness of the robot. In subsequent tests the movement of the wheels based on no of rotations from one location to the other in a stipulated time limit was tested. Then the above sequence was repeated with fast movement by reducing the time limit and again testing the fitness functional parameters of the robot.

Simulation was divided in to discrete steps translating the process from starting time to finishing time $T_s$ - $T_f$ + C where C is coarseness of simulation over a range of 0-1. Mean while other simulation parameters such as pressure applied to the physical parts of the robot was also monitored with respective sensor networks to predict the overall performance of the robot i.e. force and torque. During simulation we have monitored several parameters that play major concerns in designing the robot like position, velocity, and acceleration and so on.

The aim of our experiment is to realize the advantages in using biologically inspired design techniques in building robots in terms of hardware and software components. By simulating the robots and then monitoring several key parameters we can build robots with limited resources using the influential data. Watching out for anticipated failures in aging of components we could avoid catastrophic problems. Further this was extended to unanticipated failure cases also

### 5.1 Khepera Robot in a flat surface

To analyze the behavior of digital robot in the virtual world we have selected a flat land surface where there are no bumps and disturbances. During this test we have started with the fitness value as 0 and after 30 minutes of simulation to reach the target in a flat land with and without obstacles we achieved the fitness function value of 0.7 and the average fitness function for reaching the target by relocating the robot in other corners of the virtual world was improving to 0.9. Further if we increase the no of simulation steps above 1,000 certainly we can reach the maximum fitness level. Failure rate for sensors and joints were in the region of 2 out of 25 and out of 25 for with and without obstacles respectively.

### 5.2 Khepera Robot in a bumpy surface

In bumpy surface we have performed movement test for the Khepera robot. Since there is a penalty for each move if it does not satisfy the distance between wheels and surface. Here we had some diversified results. First with out obstacles we got a fitness value of 0.4 and by increasing them slowly over 2000 steps in the same time limit of 30 minutes it has reached the maximum fitness value of 0.6. Performance of the robot suddenly changes due to the influence of change in environmental parameters.

**Note:** - Test results shown in the table I is for target reaching test of about 25 times in two surfaces and they are only average values under a simulation time of maximum 30 minutes.

**Table I.** Comparison of Virtual Robot's Performance in different environments

| S. No | Name of the Parameter | Flat Surface | | Bumpy Surface | | Combined | |
|---|---|---|---|---|---|---|---|
| | | *Without obstacle* | *With Obstacle* | *Without obstacle* | *With Obstacle* | *Without obstacle* | *With Obstacle* |
| 1 | Fitness value | 0 to 0.9 | 0 to 0.7 | 0 to 0.6 | 0 to 0.4 | 0.6 | 0.3 |
| 2 | No of Rotations by Left wheel | 13 | 27 | 38 | 42 | 27 | 33 |
| 3 | No of Rotations by Right wheel | 14 | 23 | 36 | 44 | 39 | 42 |
| 4 | Sensor Performance | 80% | 65% | 58% | 49% | 58% | 49% |

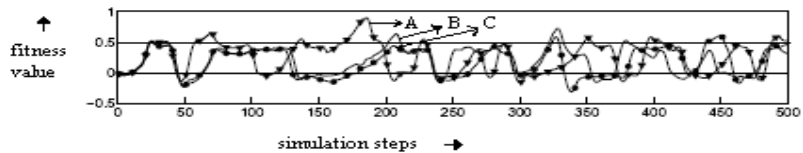

**Figure 4.** Fitness graph for different types of surfaces.

Figure 4 shows the effect of sensors in joints and angles of the robot with respect to the user defined fitness function including penalty value for keeping low distance between surface and robot's bottom section to avoid physical component failures in the virtual world A)flat B)Bumpy C)Combined. Using the above data obtained by from simulation using Framsticks. Simulation data will help selecting the right sort of hardware components like motors used in joints, sensors, control programs and even walking strategies also. Several parameters like acceleration, force, and walking strategies can be used for physical robot design which reduces considerable time. Further this could be extended to other applications like surgical robots, industrial robots, land exploring robots also.

### 5.3 Anticipated and Unanticipated Failures

To detect such failures we have to pass through controller and simulator evolution. The exploration phase (i.e. controller) evolves a controller for the physical robot using a robot simulator. The estimation phase (i.e. simulator) evolves a robot simulator, given sensor data generated by the physical robot using the controller evolved in the previous phase. Initially, the exploration phase is run, given an approximate simulation of the robot and its environment. Once terminating estimation phase, the best evolved controller can be dumped in to the physical robot. The robot then behaves, and the resulting sensor data is then supplied, along with the evolved controller, to the estimation phase. The estimation phase evolves the simulator so that the simulated robot, given the previously evolved controller, produces the same sensor data as the real robot. This helps in selecting appropriate hardware parts (sensors, wheels, etc) and software components (evolved controller programs) which will certainly avoid catastrophic failures and to have a fool proof robot design by reducing considerable amount of time.

$$F(R_i) = [\Sigma F_p(i) - F_s(i)] / C \qquad (2)$$

Fitness function for simulation phase to detect failure of both type is describe as above where Ri is the type of robot and Fp for physical robot and Fs for simulator one. C represents no of controllers evolved in reference with equation (1). The above process was repeated for n of times.

| Cases | Name of the Parameter |
|-------|----------------------|
| 1 | One of the motor weakens |
| 2 | Left wheel damage |
| 3 | Right wheel damage |
| 4 | Body damage |
| 5 | Neurons that control 2 wheels respectively |
| 6 | Any one of the Sensor fails |
| 7 | Joints fail |
| 8 | Hidden neuron fails |
| 9 | Nothing fail |

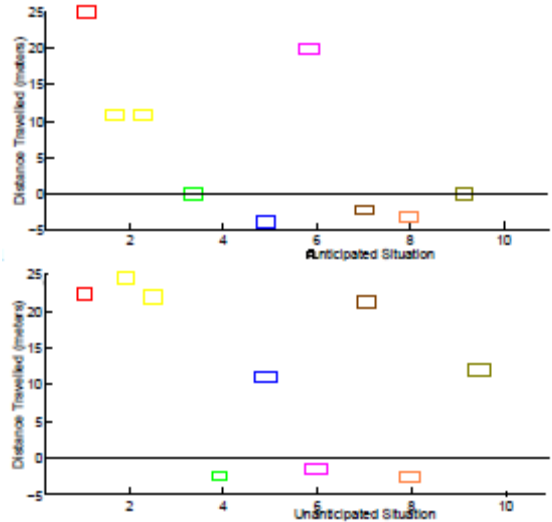

**Figure 5.** Anticipated / Unanticipated Failure distribution.

Failures categorized in both anticipated and unanticipated manners when we perform designing and testing. This data helps us in coming up with a robust structure and program design to make the robot survive in extremely critical conditions that one would be either anticipating or even a worst unanticipated form. Here neural network failure happens very rarely but where as hidden layer may fail at any time. Failure of joints is highly unpredictable. Bodily damage is very rare and happens only in unexpected cases. Sensors may fail due to the environmental or operational conditions. It is obvious that the robot's parts can be tested according to the methods suggested and can be effectively designed.

## 6   Conclusions

With the help of proposed simulation techniques we were able to capture the best design for Khepera robot and to identify complex behaviors of the same with different fitness levels. It also helps capturing the finest details / movements of the robot in a set of artificial environments. Methods suggested in this paper vindicate that effective design of robot by evolving them in an artificial environment provides sufficient information to choose minimal hardware and software parts. This improves the standard in design as well as testing robots in a successful way especially in games and automation industries i.e. forward and backward compatibility for robot design like simulation to reality and vice versa testing can be done. It also avoids the risk involved in complex robot design and provides plenty of minute information to pick individual parameters and by concentrating on low level data items when we have to deal with anticipated and unanticipated problems involved in design and testing phases with considerably less failure rate. In future this can be extended to design and test complex automation robots by predicting their failure rates to avoid wasting resources and man-hours.